\documentclass{article}

\PassOptionsToPackage{numbers, compress}{natbib}

\usepackage[english]{babel}

\usepackage[letterpaper,top=2cm,bottom=2cm,left=3cm,right=3cm,marginparwidth=1.75cm]{geometry}
\usepackage[parfill]{parskip}
\usepackage[utf8]{inputenc} 
\usepackage[T1]{fontenc}    
\usepackage{microtype}      
\usepackage{inconsolata}    

\usepackage{amsmath}        
\usepackage{amsfonts}       
\usepackage{amssymb}
\usepackage{amsthm}
\usepackage{mathtools}
\usepackage{nicefrac}       

\usepackage{graphicx}
\usepackage{booktabs}       
\usepackage{tabularx}       
\usepackage{multirow}
\usepackage{makecell}
\usepackage[table]{xcolor}  
\usepackage{float}
\usepackage{rotating}       
\usepackage{subcaption}     
\usepackage{wrapfig}
\usepackage{pifont}
\usepackage[shortlabels]{enumitem}

\usepackage{listings}
\usepackage[most]{tcolorbox}
\tcbuselibrary{listings}

\usepackage{url}            
\usepackage{natbib}
\usepackage[colorlinks=true, allcolors=blue]{hyperref}

\definecolor{promptblue}{HTML}{2B5B9E}
\definecolor{promptplum}{HTML}{8A5CD6}
\definecolor{promptolive}{HTML}{556B2F}

\lstdefinestyle{jsontool}{
  basicstyle=\scriptsize\ttfamily,
  breaklines=true,
  breakatwhitespace=true,
  showstringspaces=false,
  columns=fullflexible,
  keepspaces=true,
  stringstyle=\color{promptplum!85!black},
  keywordstyle=\color{promptblue!85!black}\bfseries,
  commentstyle=\color{gray!70!black}\itshape,
  morestring=[b]",
  morecomment=[l]{//},
  morekeywords={true,false,null},
  literate={:}{{\textcolor{black}{:}}}1,
}

\newtcblisting{toolsjson}[2][promptolive]{
  enhanced,
  colback=#1!4!white,
  colframe=#1!80!black,
  colbacktitle=#1!85!black,
  coltitle=white,
  fonttitle=\bfseries\small,
  title={#2},
  boxrule=0.6pt,
  titlerule=0pt,
  arc=3pt,
  left=8pt, right=8pt, top=6pt, bottom=6pt,
  listing only,
  listing options={style=jsontool},
}

\newcommand{\PaperBananaBench}{PaperBananaBench}

\newcommand{\mn}{\textsc{FigTree}}

\title{Figures as Programs: Recursive Generation of Editable Scientific Figures}

\author{%
  \parbox{\dimexpr\textwidth-2\tabcolsep\relax}{\centering
    Yepeng Liu$^{1,*,\dagger}$, 
    Dasen Dai$^{2,*}$, 
    Chengzhi Liu$^{1}$, 
    Yiren Song$^{3}$, 
    Hai Ci$^{3}$, 
    Yu Zhang$^{4}$, \\
    Qi Zhang$^{5}$, 
    Mike Zheng Shou$^{3}$, 
    Xin Eric Wang$^{1}$,
    Yuheng Bu$^{1}$ \\[1.4ex]
    \normalsize
    $^{1}$UC Santa Barbara \quad
    $^{2}$CUHK \quad
    $^{3}$Show Lab, National University of Singapore \\
    $^{4}$University of New South Wales \quad
    $^{5}$Tongji University \quad
  }%
}

\date{}

\begin{document}
\maketitle

{\renewcommand{\thefootnote}{\fnsymbol{footnote}}%
\footnotetext[1]{These authors contributed equally to this work.}%
\footnotetext[2]{Correspondence to: yepengliu@ucsb.edu}}

\begin{abstract}
Scientific methodology figures are essential for communicating complex methods clearly, yet creating them remains labor-intensive and typically requires multiple rounds of refinement. Recent image-generation models can synthesize visually appealing raster figures, but producing a human-satisfactory result in a single generation step remains difficult. Moreover, precise edits to raster figures are challenging for both humans and models. We formulate scientific figure generation as recursive SVG program construction and propose \mn{}, a \textit{multi-agent} system that automatically transforms a scientific paper into a structured vector figure. \mn{} grounds figure content in the source paper, decomposes a figure into a hierarchy of local regions, generates each region as a short SVG program, and assembles the resulting fragments. A render-critic refinement loop jointly inspects the rendered figure and its underlying program, enabling visual defects to be traced to specific statements and accurately repaired. We conduct extensive evaluations of \mn{} on figure quality and editability, showing that \mn{} produces high-quality figures, while also enabling more effective editing than existing raster-based methods. Our code is available at \url{https://github.com/yepengliu/FigTree}.
\end{abstract}

\section{Introduction}
\label{lab:Intro}

Automated AI Scientists demonstrate the ability to propose hypotheses, run experiments, and write papers with limited human oversight \cite{novikov2025alphaevolve,yamada2025ai,chen2025ai4research,chen2026mars,si2026towards}. High-quality methodology figures are essential to a research paper, as they often reach readers before the text. However, producing such figures still requires significant human effort.

Recent state-of-the-art image generation models, such as Nano Banana Pro \cite{nanobanana} and GPT-Image-2 \cite{gptimage}, are capable of generating scientific methodology figures directly as raster images. However, these figures are often complex, containing many interdependent components, which makes satisfactory one-shot generation challenging. The resulting images may include hallucinated elements, illegible text, or insufficient resolution. Consequently, iterative refinement is often necessary, guided by either a human or a vision-language model (VLM) critic \cite{qwen25vl,llava,gpt4o,gemini}. 

PaperBanana \cite{zhu2026paperbanana}, for example, introduces an agentic framework that iteratively refines a raster figure using feedback from a VLM critic, which translates identified issues into instructions for the image-generation model. However, natural language-guided refinement of raster images remains difficult for both parties: the critic must describe the intended modification with sufficient spatial precision, while the generator must apply it faithfully without disturbing unrelated content. Manual editing faces a similar limitation because raster images expose no explicit structure for selecting and modifying individual elements. These challenges motivate us to investigate a more editable and controllable representation for scientific methodology figures.

In this paper, we recast scientific figure generation from image generation to code generation: a Large Language Model (LLM) writes a vector graphics program (e.g., SVG) that renders the figure, providing editability and resolution independence. Prior works have demonstrated the potential of LLMs to generate code that produces various types of vector graphics \cite{automatikz,detikzify,starvector,chat2svg,omnisvg}, such as icons, illustrations, and simple diagrams. However, methodology diagrams in research papers are substantially more complex: they pack various elements and dense spatial relations into a long program. This shift from image generation to long-horizon code generation raises three challenges. (1) \textit{Generation reliability}: a whole figure corresponds to a long-horizon program with many coordinates, so errors may accumulate and surface as misalignment and overlap. (2) \textit{Spatial planning}: the model must lay out components from code alone, which demands spatial reasoning without visual feedback. (3) \textit{Verification and repair}: defects that are invisible in the code, such as occlusion or overflowing text, may surface only after rendering.

To address these challenges, we present \mn{}, a multi-agent system that generates a scientific figure as an editable SVG program through recursive decomposition. Our core insight is that a scientific figure is compositional: it forms a hierarchy of regions, each containing a few primitives. An SVG program mirrors this structure: it allows us to write each region as a subprogram independently, and then assemble them into a complete figure. This turns one hard global generation problem into many simpler local ones. For generation reliability, \mn{}~decomposes a scientific figure into a layout tree and writes a short program at each node, so no single step needs to reason about the whole figure.
For spatial planning, each child reports the space occupied by its fragment and exposes named ports for external connections. The parent uses these reports to place its children and draw the connections between them, since it is the first node with access to both endpoints. For verification and repair, a critic inspects both the rendered image, where defects surface, and the corresponding program, where each defect can be traced to a statement, allowing the generator to edit only the offending element in place.

Our contributions are as follows:
\begin{itemize}
\item We formulate editable scientific figure generation as recursive SVG program construction and propose \mn{}, a multi-agent system that integrates recursive decomposition, fragment merging, and iterative refinement.
\item For figure quality, we complement reference-based comparison with a fine-grained rubric. For editability, we introduce an iterative-editing evaluation to measure how effectively generated artifacts can be refined over multiple rounds.
\item We show that \mn{} produces high-quality figures, with particularly strong performance in hallucination control compared with existing methods. Moreover, \mn{} achieves superior editing efficiency and higher figure quality than raster-based methods under the same number of editing rounds.
\end{itemize}

\section{Related Work}
\label{sec:related_work}

Agentic systems now automate substantial parts of the research cycle, from hypothesis generation and verification \citep{novikov2025alphaevolve,chen2026mars,si2026towards,liu2026convexbench} to end-to-end paper production \citep{yamada2025ai,chen2025ai4research}. However, scientific figures remain largely outside these automated workflows and still require substantial human effort. We study scientific figure generation as a distinct research problem, with its own challenges in content grounding, spatial organization, verification, and iterative editing.

\paragraph{Raster Scientific Figure Generation.}
Image generation models \cite{zhang2025scimage} such as Nano Banana Pro \citep{nanobanana}, GPT-Image-2 \citep{gptimage}, FLUX.2 \citep{flux2}, and Qwen-Image-2.0 \citep{qwenimage2} can generate scientific figures directly as raster images. However, one-shot generation is often unreliable for scientific figures containing many components and dense spatial relationships. Agentic frameworks therefore augment image generators with planning and iterative refinement \citep{zhu2026paperbanana, zhao2026crafter, zhu2026autofigure}. These frameworks improve generation quality, but they do not change the underlying representation: the output remains a flattened raster image. Because individual components are not explicitly addressable, revisions must be conveyed indirectly through natural-language instructions to image generation models. This makes precise local edits difficult for the model and leaves the resulting figures cumbersome for humans to modify. Therefore, editability is a critical property for automated scientific figure generation.

\begin{figure}[!t]
    \centering
    \includegraphics[width=1\linewidth]{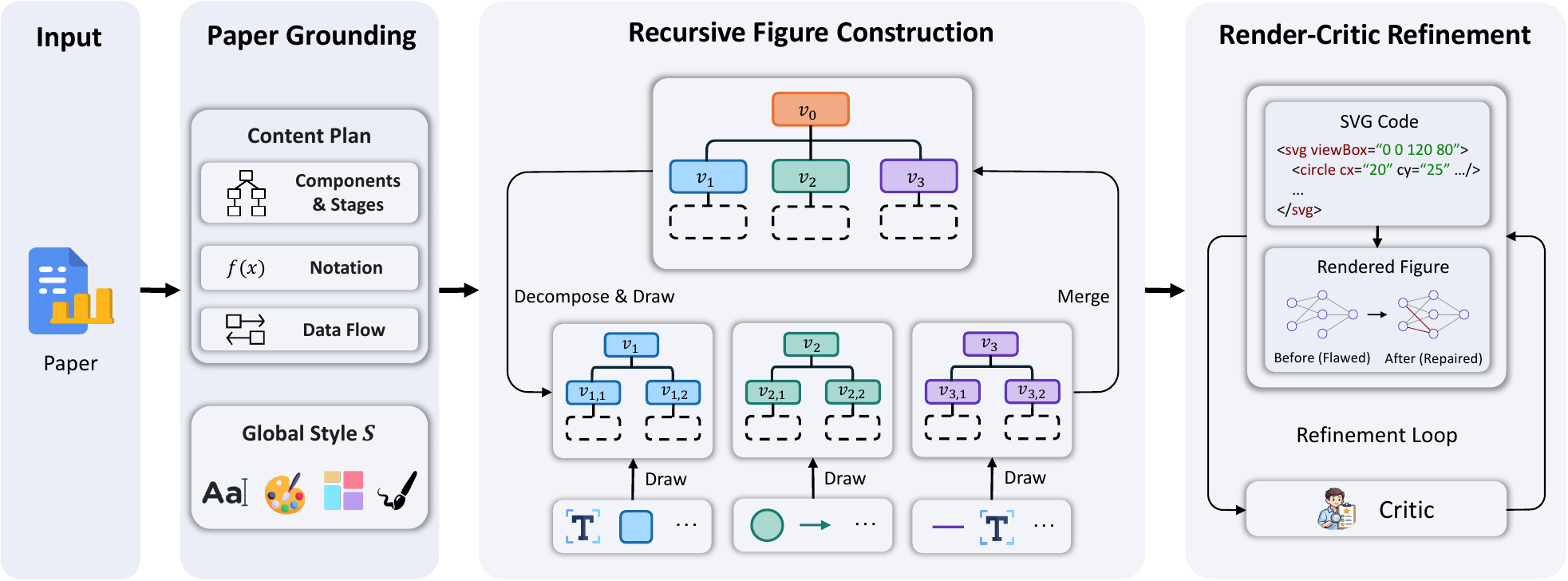}
    \caption{Workflow of \mn{}. A grounding agent extracts a content plan and global style from the input paper. Generator agents then recursively decompose the figure, draw and merge local SVG fragments, while a render-critic loop jointly inspects the SVG code and rendered output to identify and repair defects.}
    \label{fig:workflow}
\end{figure}

\paragraph{Editable Scientific Figure Generation.}

To make the generated scientific figures editable \citep{li2026automatic,liu2025presenting,ge2025autopresent,zhang2025postergen,pang2026paper2poster,tian2024chartgpt,zheng2025pptagent}, prior works \citep{zhao2026crafter,lin2026autofigureedit} adopt post-hoc vectorization pipelines that first synthesize a raster figure and then reconstruct it as an editable vector graphic through segmentation and recomposition. This approach improves the editability of the final output, but it does not address the limitations of the upstream raster generation stage, where individual components remain difficult to control or revise precisely. Moreover, the vectorization stage is optimized to reproduce the raster intermediate rather than directly recover a source-grounded figure. Therefore, errors introduced during raster generation are likely to be preserved in the final vector output. 

Another line of work generates vector graphics directly by leveraging the code-generation capabilities of LLMs \citep{automatikz,detikzify,starvector,chat2svg,omnisvg}. While this representation provides explicit structure and editability, directly producing a complete methodology figure remains challenging because the underlying graphics program can be long, and its elements are tightly coupled through spatial constraints. Concurrent work by \citet{li2026automatic} composes method illustrations by retrieving and parameterizing reusable drawing middlewares distilled from published figures, and then orchestrates them. Our work follows the direct vector generation paradigm, but addresses its long-horizon complexity through recursive program construction: we decompose a figure into localized subprograms and hierarchically generate, verify, and assemble them.

\section{Methodology}
\label{sec:methodology}

\subsection{Overview}
\label{ssec:overview}
We formulate scientific figure generation as a long-horizon SVG program construction problem. Our core idea is to leverage the hierarchical structure of a methodology diagram by generating each region as a short program fragment and progressively assembling these fragments level by level. This decomposes one long-horizon generation problem into a series of shorter, localized subproblems.

Given the text of a paper $P$, our goal is to synthesize a methodology figure that summarizes its method. We represent the figure not as an image but as an SVG program $g$, whose deterministic rendering $R(g)$ is the figure a reader sees. The task is therefore to construct a synthesizer $F$ that writes the program, $g \leftarrow F(P)$. Every visual element in the figure corresponds to an addressable statement in $g$, so a change to the figure can be implemented as a local edit to the program. This property makes the output editable, and we further exploit it during generation and repair. 

The hierarchy above is realized as a \emph{layout tree} $T=(V, E)$. Each node $v \in V$ represents a sub-figure information, specified by a tuple
\[
  v = \big(L_v,\, C_v,\, \Pi_v,\, D_v,\, S\big),
\]
where $L_v$ is a natural language description of what the sub-figure shows, $C_v=(w_v,h_v)$ is its local canvas, $\Pi_v$ is a set of boundary ports through which it connects to the rest of the figure, $D_v$ is its depth in the tree, and $S$ is the global style specification, which defines attributes such as font, palette, stroke widths, grid pitch, and arrow-marker geometry.

We instantiate $F$ as a \textit{recursive multi-agent system} (Figure~\ref{fig:workflow}): a \emph{grounder} plans the figure, a tree of \emph{generator} agents recursively draws and assembles it, and a \emph{critic} reviews every fragment. \emph{Paper grounding} (Section~\ref{ssec:grounding}) extracts from $P$ a content plan specifying what the figure must show, admitting only paper-supported content. \emph{Recursive figure construction} (Section~\ref{ssec:construction}) expands the layout tree top-down: an agent at each node either draws a leaf fragment or delegates to child agents, then merges the returned fragments. \emph{Render-critic refinement} (Section~\ref{ssec:repair}) rasterizes the program, inspects both the image and the code, and edits the offending element in place.

\subsection{Paper Grounding}
\label{ssec:grounding}

In the first stage, a \emph{grounder} extracts methodology content from the input paper and produces two artifacts. The first is a global diagram description $L_{v_0}$, which serves as a content plan specifying the components to be visualized, the data flow among them, and the notation associated with each stage. The second is a global style specification, $S$. All subsequent stages adhere to $S$, ensuring that the subfigures share a consistent visual language and that the assembled figure forms a coherent whole.

\subsection{Recursive Figure Construction}
\label{ssec:construction}

In the second stage, the figure is constructed by expanding the layout tree top-down from the root node $v_0$. Each node is processed by a \emph{generator} agent, which receives the $v$ and produces the corresponding SVG fragment.

Given $v$, the \textit{generator} makes a decision: if $v$ is atomic or its depth reaches the predefined limit $D_{\max}$, it draws $v$ as a leaf; otherwise, it decomposes $v$ into a small set of child nodes and recursively delegates each child to a new generator. Let $g_v$ denote the SVG fragment generated for node $v$, and let $\mathrm{ch}(v) = \textsc{Decompose}(v)$ denote its set of child nodes. The recursive generation process is defined as
\[
  g_v =
  \begin{cases}
    \textsc{Draw}(v), & \text{$v$ is a leaf},\\[2pt]
    \textsc{Merge}(v, \{g_c\}_{c \in \mathrm{ch}(v)}), & \text{otherwise}.
  \end{cases}
\]

We next describe the three key components of this recursion: how a node is decomposed into children ($\textsc{Decompose}$), how a leaf is drawn ($\textsc{Draw}$), and how children are assembled ($\textsc{Merge}$).

Construction proceeds by messages between a parent and its children, in two directions. Downward, a parent creates each child as a task $(L_c, C_c, \Pi_c, D_c, S)$, specifying what to draw, where to draw, the ports it exposes, its depth, and the global style. Upward, a child returns four things: its fragment $g_c$, the bounding box that the fragment occupies, exposed ports $\Pi_c$, and any remaining defects not fixed by its local repair (Section \ref{ssec:repair}). A parent places and connects its children using these reports, and a child is built without any knowledge of its siblings.

\paragraph{$\textsc{Decompose}$.}
To decompose a node, the generator interprets its description $L_v$ and creates a child node for each component of the corresponding subfigure. The decomposition is based on semantic structure rather than the geometric layout. Each child is assigned a content-adaptive subregion of the parent canvas $C_v$. In our experiments, we restrict each split to two to four children and prohibit multi-level shortcuts: a node specifies only its immediate children, while any child that remains complex is recursively decomposed at the next level. As a result, each generator produces only a short fragment containing a small number of primitives. This improves the reliability of a long-horizon task: an incorrect coordinate affects only the corresponding fragment rather than shifting all subsequently generated elements.

\paragraph{$\textsc{Draw}$.}
At a leaf, the generator generates the local SVG fragment directly. It draws only connections internal to the node, and any connection that extends beyond the node is instead represented by a named port in $\Pi_v$. These ports provide the sole interface through which sibling fragments can later connect to the node.

\paragraph{$\textsc{Merge}$.}
At an internal node, the \emph{generator} assembles its children's fragments into one using returned messages. It places each child at its assigned offset on the parent canvas and establishes the connections between them. These inter-child connections are drawn by the parent because each child is unaware of its siblings, whereas the parent has access to both endpoints and can connect them through their exposed ports. This assigns every cross-boundary connection a single owner, preventing duplicate edges that could arise if sibling nodes independently rendered their respective sides.

\subsection{Render-Critic Refinement}
\label{ssec:repair}

Decomposition makes each fragment short and easy to implement, but it does not allow the generator to verify whether the result matches its intent. Therefore, we introduce a \emph{critic} that inspects each fragment and provides feedback to the \emph{generator} for refinement.

\paragraph{Dual-space review.} Specifically, the \emph{critic} is given both the SVG fragment $g_v$ and its rendering $R(g_v)$ for inspection. These two views are complementary and enable more effective refinement. With the visual view, the critic can inspect how the elements look once rendered together. Although the generator reasons over code and can keep every statement valid, it does not see the rendered figure during generation. Some defects are therefore visible only in the rendered result, such as overflowing text or a misaligned arrow. With the code view, every element has an address, so once a defect is identified visually, the critic can trace it to the corresponding statement and provide a precise fix, such as the font size, coordinates, or color.

\paragraph{Local repair loop.} After the dual-space review, the critic returns a report containing a visual description of each defect, the anchor of the affected element, and a corresponding repair instruction. The generator uses this report to edit only the offending element in place, leaving the remainder of the fragment unchanged, and then renders the fragment again. This review-and-repair loop continues until no defect is detected or the retry budget is exhausted. Any defect that persists beyond the budget is preserved in the node's report as a residual defect rather than being silently discarded. An ancestor node, which has greater flexibility to reposition larger blocks, can then address defects that could not be resolved within the node's limited local canvas. Because the repair modifies only the affected element instead of regenerating the entire fragment, it avoids disrupting components that are already correct.

\paragraph{Verification memory.}
Rather than relying on an ad hoc and vague ``looks good'' judgment, we equip the \textit{critic} with a memory of reusable defect-repair rules. We construct this memory via an offline self-evolution stage on AutoFigure-Edit Bench~\cite{lin2026autofigureedit}. For each example, the system renders generated SVGs, prompts the critic to identify recurrent visual and structural defects, and distills them into reusable $(\text{condition}, \text{repair})$ pairs. At evaluation time, the \textit{critic} exploits the frozen rule set as an explicit checklist to inspect each fragment and apply the corresponding refinement.

\section{Experiments}
\label{sec:experiments}

We evaluate \mn{} as a system for producing methodology figures that are faithful to a paper, concise in presentation, legible at publication scale, and editable after generation. The experiments address four questions:

\begin{enumerate}
\setlength{\itemsep}{0.1pt}
\setlength{\parskip}{0pt}
\setlength{\parsep}{0pt}
    \item How does \mn{} compare with standalone image generators and agentic raster frameworks?
    \item Where do raster and vector methods differ across the diagnostic dimensions?
    \item Which components account for \mn{}'s final quality?
    \item How well do vector and raster representations improve quality under multi-turn edits?
\end{enumerate}

\subsection{Experiment Settings}
\label{ssec:exp_settings}

\paragraph{Dataset.}
Our primary test set is \PaperBananaBench{} \citep{zhu2026paperbanana}, which contains 292 methodology-figure cases curated from NeurIPS 2025 papers. Each case provides a method description, a figure caption, and an author-drawn figure. The collection covers model architectures, learning pipelines, multi-stage systems, and other schematic explanations common in machine-learning papers. At generation time, a method receives only the method description and caption. The author figure is held out and used exclusively by the reference-based evaluator. The task is therefore to reconstruct the scientific message from text, rather than trace or imitate the target image. A missing, corrupt, or policy-refused output remains in the evaluation and is counted as a loss against the author figure.

\paragraph{Baselines.}
We organize the comparison by generation procedure and output representation.

\begin{itemize}
\item \emph{Proprietary raster generator}: We evaluate Nano Banana Pro \citep{nanobanana}, GPT-Image-2 \cite{gptimage}, FLUX.2-Pro \cite{flux2}, Wan2.6-Image \citep{wan26image}, and Qwen-Image-2.0-Pro \cite{qwenimage2}. These models provide strong references for visual synthesis and surface rendering, but their outputs do not expose labels, shapes, and connectors as independently editable objects.
\item \emph{Agentic raster frameworks}: We compare with PaperBanana \citep{zhu2026paperbanana} and Crafter \citep{zhao2026crafter}. Both place planning, criticism, or iterative refinement around an image generator. 

\item \emph{Editable vector generator}: AutoFigure-Edit \citep{lin2026autofigureedit} is the published editable baseline and produces SVG output. We also construct \emph{Direct-SVG}, which prompts Claude Sonnet 4.6, the same base SVG-generation model used by \mn{}, to produce the complete figure in a single call. Unlike the full \mn{} system, Direct-SVG does not use recursive decomposition or Gemini 3.1 Pro critic-guided repair. This baseline therefore controls for the generation backbone and isolates the contributions of recursive construction and targeted repair.

\end{itemize}

\paragraph{Implementation Details.}
For methods run in our evaluation, the method text and caption are kept identical across systems. Standalone generators are queried once, while released frameworks retain their default agent graphs and iteration budgets. 

\mn{} uses Claude Sonnet 4.6 as the base model for recursive SVG construction and targeted SVG repair, and Gemini 3.1 Pro as the critic for output verification and repair guidance (full prompts in Appendix~\ref{app:methodology_prompts}). \mn{} uses an adaptive tree depth capped at $D_{\max}=10$, with two to four children at each decomposition step and a three-round repair budget. Recursion terminates earlier when a node is atomic. All nodes share one style record. The system outputs an SVG together with a corresponding figure.

\begin{table}[!ht]
\centering
\caption{Reference-based evaluation results (\%) on \PaperBananaBench{} across four quality dimensions. \textbf{Bold} and \underline{underlined} values denote the best and second-best results, respectively. \mn{} achieves the best performance across all four dimensions and improves the overall score over the strongest baseline by 9.59.}
\label{tab:paperbanana_main}
\scriptsize
\small
\setlength{\tabcolsep}{7pt}
\renewcommand{\arraystretch}{1.08}
\begin{tabular}{lrrrrr} 
\toprule
\textbf{Method}
& \textbf{Faithfulness}
& \textbf{Conciseness}
& \textbf{Readability}
& \textbf{Aesthetics}
& \textbf{Overall} \\
\midrule
\multicolumn{6}{l}{\emph{Standalone generators (proprietary)}} \\
Wan2.6-Image
  & 3.77 & 3.77 & 3.77 & 11.64 & 7.53 \\
Nano Banana Pro
  & 25.34 & 36.30 & 37.33 & 50.68 & 22.60 \\
GPT-Image-2
  & 8.56 & 4.11 & 1.71 & 40.75 & 1.37 \\
FLUX.2-Pro
  & 0.00 & 7.53 & 7.53 & 7.53 & 0.00 \\
Qwen-Image-2.0-Pro
  & 3.77 & 15.41 & 19.18 & 7.53 & 0.00 \\
\midrule
\multicolumn{6}{l}{\emph{Agentic raster frameworks}} \\
PaperBanana (w/ NB Pro)
  & 31.51 & 52.74 & 44.86 & 61.30 & 35.96 \\
Crafter (w/ NB Pro)
  & \underline{37.67}
  & \underline{55.48}
  & \underline{48.63}
  & \underline{65.41}
  & \underline{50.00} \\
\midrule
\multicolumn{6}{l}{\emph{Editable and structured systems}} \\
AutoFigure-Edit (w/ NB2)
  & 12.33 & 6.85 & 3.42 & 7.88 & 1.37 \\
Direct-SVG
  & 30.82 & 27.05 & 7.53 & 15.41 & 3.77 \\
\rowcolor{blue!8}
\textbf{\mn{}}
  & \textbf{46.23}
  & \textbf{64.38}
  & \textbf{51.37}
  & \textbf{66.78}
  & \textbf{59.59} \\
\bottomrule
\end{tabular}
\vspace{0.5em}
\end{table}

\paragraph{Evaluation.} We evaluate each method on the following three protocols (full prompt templates in Appendix~\ref{app:prompts}):

\begin{itemize}

  \item \emph{Reference-based evaluation.} We follow the \PaperBananaBench{}. Given the same methodology, text, and caption, a VLM compares each generated figure against the corresponding held-out author-created figure and independently assesses four dimensions: \emph{faithfulness}, \emph{conciseness}, \emph{readability}, and \emph{aesthetics}. Faithfulness assesses technical alignment with the source material, including logical flow, scope, and the absence of hallucinated content. Conciseness evaluates whether the figure provides an effective abstraction rather than a box-by-box transcription of the paper. Readability captures legibility, edge routing, element overlap, visual contrast, and effective use of the canvas. Aesthetics assesses visual hierarchy, typography, color harmony, and overall publication-quality polish.

Each comparison produces one of three outcomes: \emph{Model}, \emph{Tie}, or \emph{Human}. Model wins, ties, and human wins are assigned weights of (1), (0.5), and (0), respectively. Let $n_d^{\mathrm{M}}$ and $n_d^{\mathrm{T}}$ denote the numbers of model wins and ties for dimension $d$, and $N$ the total number of cases. The reported score is
\begin{equation}
S_d=100\,\frac{n_d^{\mathrm{M}}+0.5n_d^{\mathrm{T}}}{N}.
  \label{eq:pb_win_rate}
\end{equation}

The overall score is evaluated using a two-tier approach. Faithfulness and readability form the first tier; if they identify a clear winner, that decision becomes the overall outcome for the case. Otherwise, conciseness and aesthetics are used as a second-tier tie-breaker. The resulting per-case outcomes are then aggregated using Eq.~\eqref{eq:pb_win_rate}.

\item \emph{Fine-grained diagnostic evaluation.}
Reference-based evaluation measures preference relative to the author-created figure but does not identify specific errors. We therefore introduce an absolute 1-10 rubric comprising eight diagnostic dimensions: (1) content grounding, including \emph{semantic coverage} and \emph{hallucination control}; (2) diagram structure, including \emph{topology accuracy} and \emph{flow readability}; (3) rendering quality, including \emph{text/glyph quality} and \emph{visual polish}; and (4) publication readiness, including \emph{academic style} and \emph{compactness}. The judge is provided with the source text, caption, reference figure, and generated output, and uses the source text as the authoritative basis for evaluating semantic correctness. The overall score is computed as the mean of the eight-dimensional scores.

\item \emph{Iterative-editing evaluation.}
We construct two edit branches from the same defect-injected figure: the vector branch retains the source SVG, and the raster branch receives its
rendering. Over eight editing rounds, a VLM proposes one edit request per branch per round based on the defects remaining in that branch, and the two branches share the same initial request. After each round, a judge evaluates the quality of both branches. The quality score is evaluated with fine-grained diagnostic evaluation (Section~\ref{ssec:exp_editability}).

\end{itemize}

\begin{table}[t]
\centering
\caption{Fine-grained diagnostic evaluation on \PaperBananaBench{}. Scores are on a 1--10 scale. \mn{} achieves the highest overall score, leading in hallucination control, visual polish, and academic style while matching the best result in semantic coverage.}
\label{tab:paperbanana_diagnostic}
\small
\setlength{\tabcolsep}{3pt}
\renewcommand{\arraystretch}{1.1}
\resizebox{\linewidth}{!}{%
\begin{tabular}{lccccccccc}
\toprule
\textbf{Method} 
  & \multicolumn{2}{c}{\textbf{Content Grounding}}
  & \multicolumn{2}{c}{\textbf{Diagram Structure}}
  & \multicolumn{2}{c}{\textbf{Rendering Quality}}
  & \multicolumn{2}{c}{\textbf{Publication Readiness}}
  & \multirow{2}{*}{\textbf{Overall}} \\
\cmidrule(lr){2-3} \cmidrule(lr){4-5} \cmidrule(lr){6-7} \cmidrule(lr){8-9}
& \textbf{Sem. Cov.} & \textbf{Halluc. Ctrl.}
& \textbf{Topology} & \textbf{Flow}
& \textbf{Text/Glyph} & \textbf{Polish}
& \textbf{Acad. Style} & \textbf{Compactness} & \\
\midrule
\multicolumn{10}{l}{\emph{Standalone generators (proprietary)}} \\
Wan2.6-Image
  & 3.90 & 3.95 & 4.24 & 5.76 & 3.67 & 4.48 & 3.33 & 7.19 & 4.57 \\
Nano Banana Pro
  & 6.41 & 3.36 & 5.36 & 5.95 & 5.77 & 4.05 & 2.91 & 6.50 & 5.04 \\
FLUX.2-Pro
  & 4.24 & 3.81 & 4.24 & 5.43 & 3.52 & 5.10 & 3.86 & 7.05 & 4.66 \\
Qwen-Image-2.0-Pro
  & 3.85 & 4.95 & 4.00 & 6.85 & 8.00 & 6.65 & 4.70 & 7.90 & 5.86 \\
GPT-Image-2
  & \textbf{8.41} & \underline{7.45} & \textbf{8.50} & \textbf{8.26}
  & \underline{8.36} & \underline{8.32} & \underline{8.86}
  & \textbf{8.05} & \underline{8.28} \\
\midrule
\multicolumn{10}{l}{\emph{Agentic raster frameworks}} \\
Crafter
  & \underline{7.00} & 6.18 & 6.91 & \underline{8.23}
  & \textbf{8.55} & 8.00 & 7.41 & 7.05 & 7.42 \\
PaperBanana
  & 6.55 & 4.73 & 6.05 & 6.73 & 6.64 & 5.95 & 4.95 & 6.95 & 6.07 \\
\midrule
\multicolumn{10}{l}{\emph{Editable and structured systems}} \\
Direct-SVG
  & 6.45 & 6.68 & 6.05 & 6.82 & 6.36 & 6.36 & 6.23 & 6.23 & 6.40 \\
AutoFigure-Edit
  & 1.95 & 1.55 & 1.50 & 1.64 & 2.32 & 1.77 & 1.14 & 1.64 & 1.69 \\
\rowcolor{blue!8}
\textbf{\mn{}}
  & \textbf{8.41} & \textbf{8.59} & \underline{8.14} & 8.13
  & \underline{8.36} & \textbf{8.64} & \textbf{8.91}
  & \underline{7.95} & \textbf{8.39} \\
\bottomrule
\end{tabular}
}
\end{table}

\subsection{Main Results}
\label{ssec:exp_main_results}

\paragraph{Reference-based evaluation.} Table~\ref{tab:paperbanana_main} presents the reference-based evaluation results. \mn{} achieves the strongest performance across four dimensions. Its overall score exceeds Crafter by 9.59 points and PaperBanana by 23.63 points. Compared with the strongest baseline, Crafter, our method achieves its largest gains in faithfulness (+8.56) and conciseness (+8.90), along with more modest improvements in readability (+2.74) and aesthetics (+1.37). This indicates that our method effectively controls the content hallucination and produces a more concise abstraction of the method, while remaining competitive in visual presentation.

\paragraph{Fine-grained diagnostic evaluation.}
Table~\ref{tab:paperbanana_diagnostic} provides a fine-grained analysis across eight diagnostic dimensions. Compared with the baselines, \mn{} achieves the highest overall score and ranks first or ties for first on four of the eight dimensions: hallucination control, semantic coverage, visual polish, and academic style. On each of the remaining four dimensions, it stays within 0.4 points of the best method. Specifically, \mn{} achieves its largest advantage in hallucination control while remaining competitive on the other dimensions. These results further suggest that the vector representation and recursive construction are particularly effective at preserving semantic correctness without sacrificing visual polish or publication readiness. Additional qualitative examples and a cost--quality analysis are provided in Appendix~\ref{sec:appendix_additional_results}.

\subsection{Ablation Study}
\label{ssec:exp_ablation}

We ablate five core components of \mn{}: paper grounding, recursive decomposition, parent-owned cross-boundary connections, render-critic refinement, and verification memory. Across all variants, we keep the source text, style specification, generation backend, and evaluator fixed.

\begin{table}[h]
\centering
\caption{Mechanism study on a sampled benchmark subset. Scores are on a 1--10 scale; $\Delta$ values indicate the difference from full \mn{}. Full \mn{} achieves the highest overall score, while removing verification memory causes the largest degradation, followed by paper grounding and recursive decomposition.}
\label{tab:TreeSVG_ablation}
\scriptsize
\setlength{\tabcolsep}{6pt}
\renewcommand{\arraystretch}{1.1}
\begin{tabular}{l r r r r r}
\toprule
\textbf{Configuration} & \textbf{Content} & \textbf{Structure} & \textbf{Rendering} & \textbf{Pub. Ready} & \textbf{Overall} \\
\midrule
\rowcolor{blue!5}
\textbf{\mn{}} & \textbf{8.50} & \textbf{8.18} & \textbf{8.44} & \textbf{8.39} & \textbf{8.38} \\
\midrule
\multicolumn{6}{l}{\textit{Ablation variants}} \\
\quad \textit{w/o Render-Critic Refinement} & 8.34 & 7.80 & 7.55 & 7.75 & 7.86 \\
\rowcolor{gray!10} \quad $\Delta$ & \textcolor[rgb]{0,0.5,0}{$-0.16$} & \textcolor[rgb]{0,0.5,0}{$-0.38$} & \textcolor[rgb]{0,0.5,0}{$-0.89$} & \textcolor[rgb]{0,0.5,0}{$-0.64$} & \textbf{\textcolor[rgb]{0,0.5,0}{$-0.52$}} \\
\addlinespace[2pt]
\quad \textit{w/o Parent-owned Connections} & 7.84 & 6.80 & 6.75 & 6.71 & 7.03 \\
\rowcolor{gray!10} \quad $\Delta$ & \textcolor[rgb]{0,0.5,0}{$-0.66$} & \textcolor[rgb]{0,0.5,0}{$-1.38$} & \textcolor[rgb]{0,0.5,0}{$-1.69$} & \textcolor[rgb]{0,0.5,0}{$-1.68$} & \textbf{\textcolor[rgb]{0,0.5,0}{$-1.35$}} \\
\addlinespace[2pt]
\quad \textit{w/o Paper Grounding} & 4.98 & 5.30 & 7.09 & 6.30 & 5.91 \\
\rowcolor{gray!10} \quad $\Delta$ & \textcolor[rgb]{0,0.5,0}{$-3.52$} & \textcolor[rgb]{0,0.5,0}{$-2.88$} & \textcolor[rgb]{0,0.5,0}{$-1.35$} & \textcolor[rgb]{0,0.5,0}{$-2.09$} & \textbf{\textcolor[rgb]{0,0.5,0}{$-2.47$}} \\
\addlinespace[2pt]
\quad \textit{w/o Recursive Decomposition} & 5.05 & 5.57 & 7.66 & 6.80 & 6.27 \\
\rowcolor{gray!10} \quad $\Delta$ & \textcolor[rgb]{0,0.5,0}{$-3.45$} & \textcolor[rgb]{0,0.5,0}{$-2.61$} & \textcolor[rgb]{0,0.5,0}{$-0.78$} & \textcolor[rgb]{0,0.5,0}{$-1.59$} & \textbf{\textcolor[rgb]{0,0.5,0}{$-2.11$}} \\
\addlinespace[2pt]
\quad \textit{w/o Verification Memory} & 3.27 & 3.93 & 5.57 & 4.95 & 4.43 \\
\rowcolor{gray!10} \quad $\Delta$ & \textcolor[rgb]{0,0.5,0}{$-5.23$} & \textcolor[rgb]{0,0.5,0}{$-4.25$} & \textcolor[rgb]{0,0.5,0}{$-2.87$} & \textcolor[rgb]{0,0.5,0}{$-3.44$} & \textbf{\textcolor[rgb]{0,0.5,0}{$-3.95$}} \\
\bottomrule
\end{tabular}
\end{table}

The full \mn{} obtains the best overall score of 8.38. Removing verification memory causes the largest degradation, reducing the overall score to 4.43 ($-3.95$). The drop across all dimensions, especially content and structure, shows that the verification rules help the critic detect and repair recurring errors. Removing paper grounding also substantially degrades performance, lowering the overall score to 5.91 ($-2.47$). The largest drop is in content grounding, suggesting that generating directly from the paper text without an explicit content plan makes it harder to preserve paper-specific mechanisms and avoid unsupported content. Removing recursive decomposition lowers the overall score to 6.27 ($-2.11$). While rendering quality remains relatively strong, the substantial drops in content and structure show that generating the entire SVG in a single long-horizon process makes it harder to preserve the method's components and their relationships.

\begin{figure}[h]
  \centering
  \includegraphics[width=0.65\linewidth]{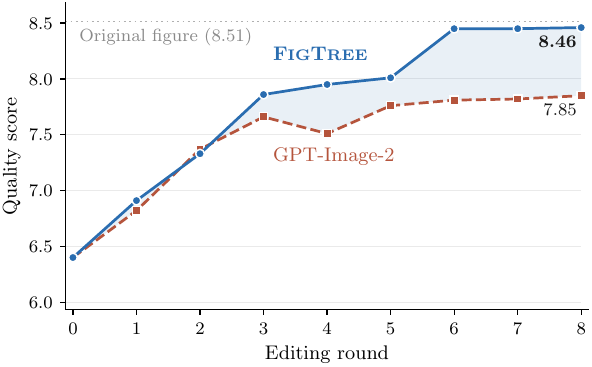}
  \caption{Quality scores across iterative editing rounds. The gray dotted line marks the quality of the original figure before defect injection. As the number of editing rounds increases, \mn{} achieves higher quality than the raster branch, demonstrating the advantages of vector-based representations in iterative refinement.}
  \label{fig:iterative_editability}
\end{figure}

\subsection{Iterative-editing Evaluation}
\label{ssec:exp_editability}
In practice, a scientific figure is rarely perfected in a single generation step and often requires multiple rounds of refinement. We therefore design an iterative editing evaluation to compare how well vector and raster representations preserve and improve figure quality over successive edits. Starting from a high-quality scientific figure, we manually introduce a set of defects to create a common defective version. A VLM is then given the original figure as the reference and target, together with the defective figure as input, and generates an edit request aimed at correcting the identified defects. 

Both branches begin from the same defective figure and receive the same initial edit request. After each editing round, the VLM examines the updated output of each branch and generates the next edit request based on the remaining defects. Figure quality at each round is measured as the mean of the eight fine-grained diagnostic scores. Figure~\ref{fig:iterative_editability} shows how figure quality evolves over eight rounds of iterative editing. As the number of editing rounds increases, \mn{} exhibits a steady improvement in quality and outperforms the raster branch. These results demonstrate that the vector-based representation enables more reliable and controllable refinement across successive edits.

\section{Conclusion}
\label{lab:Conclusion}

We presented \mn{}, a multi-agent system that formulates editable scientific figure generation as recursive SVG program construction. \mn{} grounds figure content in the source paper, decomposes complex diagrams into localized subprograms, and assembles them through hierarchical generation with parent-owned connections. A render--critic loop further enables visual defects to be traced to specific SVG elements and repaired locally. Experiments show that \mn{} produces high-quality methodology figures and supports more reliable iterative editing than raster-based approaches.

\clearpage

\bibliographystyle{unsrtnat}
\bibliography{references}

\clearpage


\appendix

\section{Additional Results and Analysis}
\label{sec:appendix_additional_results}

\subsection{Cost--Quality Trade-off}
\label{ssec:cost_quality_tradeoff}

\begin{figure}[!ht]
    \centering
    \includegraphics[width=0.8\linewidth]{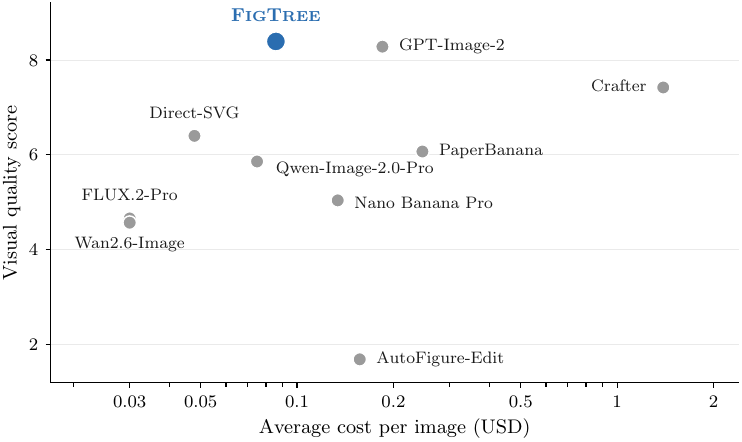}
    \caption{
    Cost--quality comparison across figure-generation methods. The
    x-axis shows the average cost per generated image in USD (lower is
    better), and the y-axis shows the visual quality score (higher is
    better). \mn{} achieves the strongest overall trade-off, combining
    the highest visual quality with relatively low generation cost.
    }
    \label{fig:cost_quality_tradeoff}
\end{figure}

In addition to the quantitative evaluation in
Section~\ref{sec:experiments}, we analyze the trade-off
between generation cost and visual quality. As shown in
Figure~\ref{fig:cost_quality_tradeoff}, \mn{} attains the highest
visual quality among the compared systems, at a lower cost per image
than GPT-Image-2 and Crafter.

This result is practically important for scientific figure generation,
where iterative prompting and repeated edits are common. A method that
offers both strong visual quality and low average generation cost is
better suited for real-world authoring workflows, especially when users
need to refine figures over multiple rounds.

\subsection{Additional Qualitative Examples}
\label{ssec:additional_qualitative_examples}

\begin{figure}[!t]
    \centering
    \includegraphics[width=\linewidth]{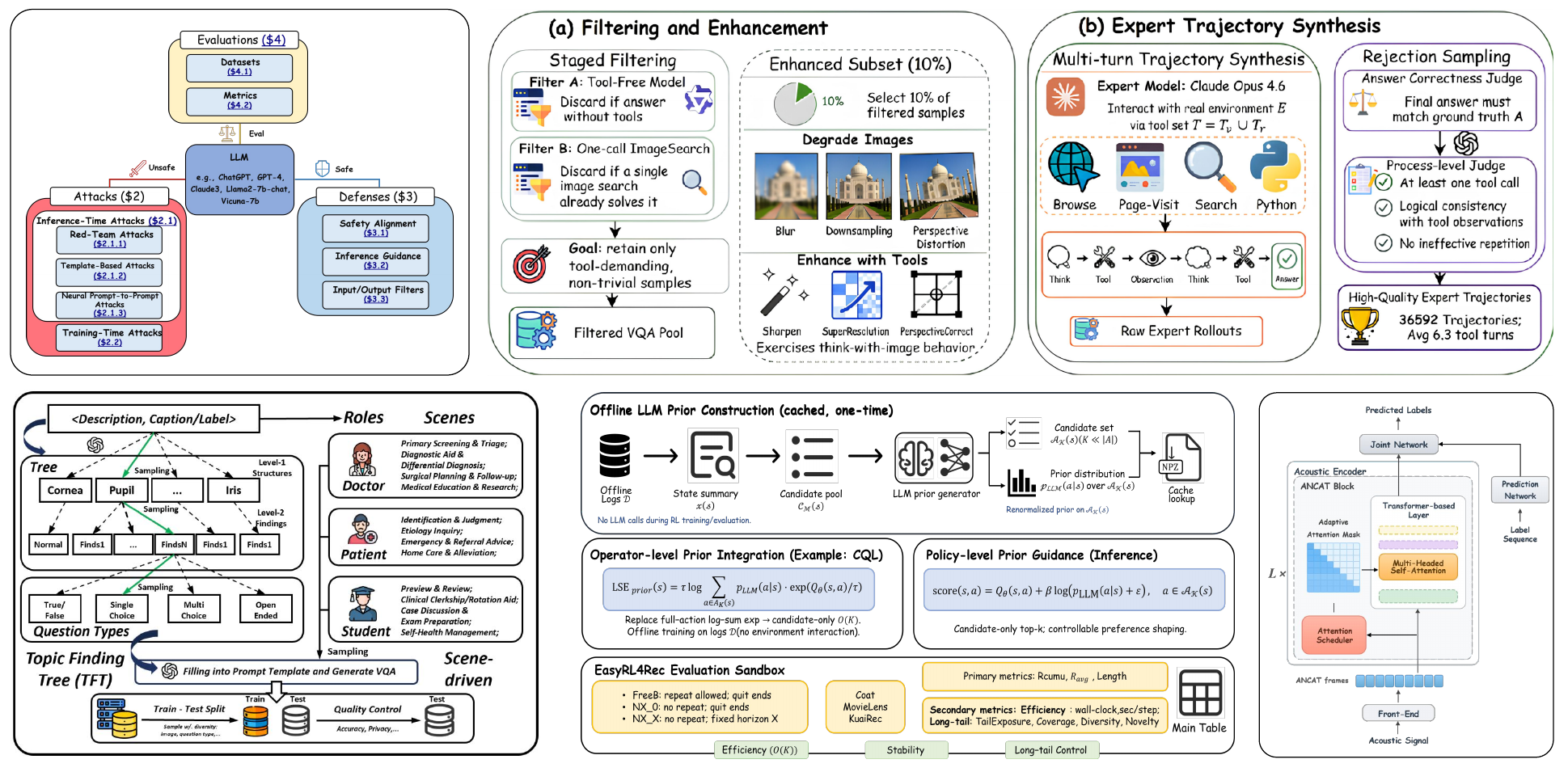}
    \caption{
    Additional qualitative examples generated by \mn{}. The examples
    cover diverse methodology-figure structures, including cyclic
    interaction workflows, neural network architectures, multi-stage
    generation pipelines, hierarchical taxonomies, equation-augmented
    optimization frameworks, and parallel-branch representation
    models. The figures are shown as rendered previews, while the
    underlying outputs are retained as structured and editable SVGs.
    }
    \label{fig:additional_demos}
\end{figure}

Figure~\ref{fig:additional_demos} presents additional methodology
figures generated by \mn{}. These examples complement the quantitative
evaluation by illustrating the diversity of scientific content and
diagram structures supported by our SVG-generation agent. The examples
span cyclic interaction workflows, neural network architectures,
multi-stage generation pipelines, hierarchical taxonomies,
equation-augmented optimization frameworks, and parallel-branch
representation models.

These methods require substantially different forms of visual
organization. Some are naturally expressed as sequential pipelines,
whereas others involve cyclic interactions, nested subsystems,
parallel computational branches, or hierarchical category structures.
Rather than mapping all methods to a fixed template, \mn{} adapts its
recursive decomposition, layout strategy, and connection structure to
the logical organization of each method. It can also integrate
heterogeneous visual elements---including text labels, containers,
connectors, mathematical expressions, neural network blocks, legends,
and small statistical graphics---within a unified visual hierarchy.

The examples in Figure~\ref{fig:additional_demos} are shown as rendered
previews for compact presentation, while their underlying outputs are
retained as structured SVGs in which labels, shapes, groups, and
connectors remain independently editable. Together, these qualitative
examples illustrate the breadth of figure types supported by \mn{},
while the quantitative experiments in
Section~\ref{sec:experiments} provide the formal evaluation of
generation quality and iterative editability.

\newpage
\section{Evaluation Prompt Templates}
\label{app:prompts}
This appendix provides the complete prompts used in the benchmark evaluation. Fields in square brackets are replaced for each example.
Images are supplied as separate multimodal inputs rather than serialized into
the text prompt.

\subsection{PaperBananaBench Evaluation Prompts}
\label{app:paperbanana_prompt}
For the PaperBananaBench reference-based protocol, the judge receives the
method section and caption where required, together with a human-drawn reference
and the model diagram. The benchmark runs four independent pairwise judgments.
Each prompt returns one of \texttt{Model}, \texttt{Human}, \texttt{Both are good},
or \texttt{Both are bad}; the reported percentage is computed from these
outcomes according to the official evaluation implementation.
The multimodal user-message templates vary slightly by dimension: for
faithfulness and conciseness, the input includes the methodology section, the
figure caption, and both the human-drawn and model-generated images; for
readability and aesthetics, only the caption and the two images are provided.
Images are attached as separate multimodal inputs, not serialized into text.
We use the standard PaperBananaBench system prompts for the four evaluation
axes. Below we summarize their core definitions, veto rules, and decision
criteria.

\paragraph{Faithfulness.}
Faithfulness measures the technical alignment between the diagram and the
paper's content. A faithful diagram must be factually correct, logically sound,
and strictly adhere to the scope defined by the caption, while preserving the
core logic flow and module interactions described in the method section.
Simplification (e.g., representing a standard module as a single block) is
encouraged, but every visual element must have a direct, non-contradictory basis
in the text. The veto rules flag major hallucinations, logical contradictions,
scope violations, and gibberish content (e.g., broken mathematical notation).
The judge selects the better option based solely on these criteria; if both
diagrams successfully meet the definition without veto errors, the outcome is
``Both are good.''

\paragraph{Conciseness.}
Conciseness is defined as the visual signal-to-noise ratio. A concise diagram
acts as a high-level abstraction, distilling complex logic into clean blocks,
flowcharts, or icons, relying on structural shorthand (arrows, grouping) and
keywords rather than lengthy textual descriptions. Veto rules prohibit textual
overload (full sentences or more than 15 words per box, except for data
examples), literal copying of the method text, and cluttering with raw
equations. The comparison follows the same four-option output scheme, with
``Both are good'' as the default when both diagrams achieve effective abstraction
without veto violations.

\paragraph{Readability.}
Readability assesses how easily a reader can extract and navigate the core
information. A readable diagram must have clear visual flow, high legibility,
and minimal interference. This dimension is treated as a pass/fail baseline:
only severe violations trigger a failure. The veto rules cover visual noise
(e.g., embedded figure titles or watermarks), occlusion and overlap, chaotic
routing, illegible font sizes, low contrast, inefficient non-rectangular
layouts (which waste space in \LaTeX{} documents), and the use of a black
background. If neither diagram violates any veto rule, the judge \emph{must}
default to ``Both are good''; winner selection is reserved for cases with clear
and substantial differences.

\paragraph{Aesthetics.}
Aesthetics refers to the visual polish, professional maturity, and design
harmony of the diagram, aiming for the standards of top-tier conferences such as
NeurIPS or CVPR. Criteria include a refined visual hierarchy, balanced white
space, consistent typography, and a harmonious color palette. Veto rules flag
low-quality artifacts (background grids, pixelation), jarring neon colors,
amateurish styling (overly rounded or clip-art-like elements), inconsistent
typography, and black backgrounds. As with the other dimensions, the output can
be ``Model'', ``Human'', ``Both are good'', or ``Both are bad''; the judge is
instructed not to force a winner when both diagrams are aesthetically
acceptable.
For all four axes, the judge must provide a structured reasoning string and
return the decision in a strict JSON format, as specified in the official
benchmark implementation.

\subsection{Fine-grained Diagnostic Evaluation Prompts}
\label{app:diagnostic_prompt}
Our diagnostic evaluator receives the reference image, the generated output,
and the source-text excerpt. The system prompt
(Figure~\ref{fig:diagnostic-system-prompt}) fixes the judging behavior. The
user prompt includes all eight dimensions, their scoring anchors, and the
required structured response. The source text is truncated to 2,000 characters
by the evaluator before insertion.

\begin{figure}[!ht]
\centering
\footnotesize
\begin{toolsjson}[promptblue]{Diagnostic Evaluator: System Prompt}
You are an expert evaluator for academic figure generation quality.
Your task is to rate a generated figure on a 1-10 scale.
You will receive:
- A **Reference** image: one possible visual realization of the figure (for layout/style reference)
- **Source Text**: the paper excerpt the figure should represent (the PRIMARY standard for content)
- One **Generated Output**: the figure to evaluate
For each dimension, you MUST:
1. Provide a score from 1-10 (integer)
2. Give a brief, specific justification (1-2 sentences) citing concrete visual evidence
3. List 1-4 concise diagnostic tags for the dominant issues or strengths
CRITICAL RULES:
- Use the FULL 1-10 range. Do NOT cluster all scores around 7-8.
- A score of 5 means "average/mediocre" --- use it.
- A score of 1-2 means "completely broken" --- use it when appropriate.
- A score of 9-10 means "near-perfect" --- reserve for truly excellent outputs.
- Judge the OUTPUT AS RENDERED --- what you actually see matters.
- For Semantic Coverage: the source text is the PRIMARY truth. The reference image is the target visual realization.
- For Hallucination Control: compare against BOTH the reference image AND the source text. Flag anything absent from both.
- For Text \& Glyph Quality: inspect INDIVIDUAL GLYPHS closely. Penalize asymmetric strokes, broken curves, blurry edges, and malformed characters even if the word is still readable.
- For Academic Suitability: prefer clean, editable, vector-like academic diagrams over photorealistic or marketing-style raster art when both communicate similar content.
- Do not give high scores for attractive figures that are semantically wrong, hallucinated, or hard to edit.
- Be strict but fair. Academic publishing standards are high.
- Respond ONLY with valid JSON, no other text.
\end{toolsjson}
\caption{System prompt used by the fine-grained diagnostic evaluator. The evaluator judges one generated figure against a reference image and the source-text excerpt, and returns a strict JSON response with per-dimension scores, justifications, and diagnostic tags.}
\label{fig:diagnostic-system-prompt}
\end{figure}

We split the full user prompt into five logical blocks: one for the overall context and source text (Figure~\ref{fig:diagnostic-user-context}), and four blocks corresponding to the dimension groups used in our evaluation table (Table~\ref{tab:paperbanana_diagnostic}): content grounding (Figure~\ref{fig:diagnostic-user-content}), diagram structure (Figure~\ref{fig:diagnostic-user-structure}), rendering quality (Figure~\ref{fig:diagnostic-user-rendering}), and publication readiness (Figure~\ref{fig:diagnostic-user-output}). Each block contains the scoring anchors and instructions for its group. The final block also includes the required JSON output schema.

\begin{figure}[!ht]
\centering
\footnotesize
\begin{toolsjson}[promptblue]{Diagnostic Evaluator: Context \& Source Text}
**Source Text** (this is the PRIMARY ground truth for what the figure must represent):

---BEGIN SOURCE---
[SOURCE_TEXT]
---END SOURCE---

---

**IMPORTANT CONTEXT:**
- The Reference Image shown above is ONE possible visual realization - use it for layout/style guidance.
- But the SOURCE TEXT is the authoritative standard for Content Fidelity: the figure must faithfully represent what the text describes.
- For Hallucination Control: any element absent from BOTH the source text AND the reference image is a hallucination.
- For Text \& Glyph Quality: zoom in and examine INDIVIDUAL GLYPHS. Look for asymmetric strokes (e.g., 'T' with unequal arms), broken curves, blurry edges, inconsistent stroke widths within a single letter. Penalize these even if the word remains readable.
- For Structural Accuracy: inspect arrow endpoints, flow direction, grouping, containment, overlaps, cropping, and whether the visual hierarchy matches the intended scientific logic.
- For Publication Readiness: evaluate whether the figure is appropriate for a conference or journal paper and uses its canvas economically, not just whether it is visually pleasing.
\end{toolsjson}
\caption{User prompt for the fine-grained diagnostic evaluator: overall context and source-text instructions. This block provides the primary ground truth and general judging guidelines that apply to all evaluation dimensions.}
\label{fig:diagnostic-user-context}
\end{figure}

\begin{figure}[!ht]
\centering
\footnotesize
\begin{toolsjson}[promptblue]{Diagnostic Evaluator: Content Grounding}
### Semantic Coverage
Does the figure cover the paper-specific concepts that matter? Check the main scientific claim, named modules/entities, algorithmic stages, inputs/outputs, and important dependencies described in the SOURCE TEXT. The reference image is the visual target, but the source text is the authority for what the figure should mean.
    Scoring anchors:
       1-2: Wrong topic or generic diagram; almost none of the paper-specific content is present
       3-4: Broad domain is recognizable, but most key modules or claims are missing
       5-6: Main idea appears, but several important mechanisms or relationships are omitted/distorted
       7-8: Most key concepts and relationships are represented with minor omissions
       9-10: Complete, paper-specific coverage of the essential concepts, stages, and relationships

### Hallucination Control
Are visible elements justified by the source text or reference figure? Penalize invented modules, fabricated labels, unsupported datasets, fictional metrics, arbitrary icons, extra branches, or decorative scenes that change the method. Minor abstraction is acceptable only when it clarifies documented content.
    Scoring anchors:
       1-2: Most content is fabricated or unrelated
       3-4: Major unsupported components or labels materially alter the method
       5-6: Core idea is recognizable, but several invented details are present
       7-8: Only minor unsupported embellishments or one small wrong label
       9-10: No meaningful hallucination; all important elements are source/reference-grounded
\end{toolsjson}
\caption{User prompt for the fine-grained diagnostic evaluator: Content Grounding group. This block contains the scoring anchors for semantic coverage and hallucination control.}
\label{fig:diagnostic-user-content}
\end{figure}

\begin{figure}[!ht]
\centering
\footnotesize
\begin{toolsjson}[promptblue]{Diagnostic Evaluator: Diagram Structure}
### Topology Accuracy
Does the structural organization match the intended method? Check hierarchy, grouping, module containment, repeated blocks, branching/merging, spatial relationships, and whether the diagram topology reflects the reference/source logic rather than a generic layout.
    Scoring anchors:
       1-2: No coherent topology; grouping and hierarchy are unusable
       3-4: Severe structural mismatch; important blocks are misplaced or incorrectly grouped
       5-6: Recognizable organization, but several hierarchy/grouping errors remain
       7-8: Mostly correct topology with minor grouping or proportional issues
       9-10: Precise, publication-grade topology matching the scientific structure

### Flow Readability
Can a reader follow the information/process flow? Check arrow direction, arrow endpoints, crossings, route clarity, reading order, alignment, spacing, overlaps, cropping, and whether the visual path explains the method without ambiguity.
    Scoring anchors:
       1-2: Flow is unreadable; arrows/paths are missing, chaotic, or misleading
       3-4: Major flow confusion from wrong arrows, crossings, overlaps, or poor reading order
       5-6: Overall flow can be inferred, but routing/spacing defects slow comprehension
       7-8: Clear flow with only minor arrow or spacing imperfections
       9-10: Very clear, economical flow with precise routing and no ambiguity
\end{toolsjson}
\caption{User prompt for the fine-grained diagnostic evaluator: Diagram Structure group. This block covers topology accuracy and flow readability.}
\label{fig:diagnostic-user-structure}
\end{figure}

\begin{figure}[!ht]
\centering
\footnotesize
\begin{toolsjson}[promptblue]{Diagnostic Evaluator: Rendering Quality}
### Text \& Glyph Quality
Judge label correctness and typographic rendering. Check spelling, terminology, notation, label placement, font consistency, paper-column readability, and individual glyph integrity. Penalize malformed raster text: asymmetric strokes, broken curves, partial letters, blurry edges, or inconsistent stroke widths.
    Scoring anchors:
       1-2: Text is illegible, garbled, or mostly missing
       3-4: Severe text errors or widespread glyph deformation
       5-6: Readable but with noticeable spelling, placement, blur, or glyph defects
       7-8: Most text is correct and clean, with minor local defects
       9-10: All labels are accurate, sharp, well placed, and typographically consistent

### Visual Polish
Judge professional visual execution: restrained palette, line sharpness, consistent stroke weights, balanced whitespace, contrast, clean containers, and absence of low-resolution artifacts. Penalize stock-like imagery, excessive gradients, noisy textures, cartoonish decoration, and clutter.
    Scoring anchors:
       1-2: Visually broken or unprofessional
       3-4: Messy composition, harsh/inconsistent styling, or obvious artifacts
       5-6: Acceptable but visibly rough or unbalanced
       7-8: Polished academic visual style with minor imperfections
       9-10: Highly polished and visually balanced; ready for formal presentation
\end{toolsjson}
\caption{User prompt for the fine-grained diagnostic evaluator: Rendering Quality group. This block contains the scoring anchors for text/glyph quality and visual polish.}
\label{fig:diagnostic-user-rendering}
\end{figure}

\begin{figure}[!ht]
\centering
\scriptsize
\setlength{\abovecaptionskip}{0pt}
\setlength{\belowcaptionskip}{0pt}
\begin{toolsjson}[promptblue]{Diagnostic Evaluator: Publication Readiness \& Output}
### Academic Style
Would the figure look appropriate in a top-tier paper? Check scientific seriousness, compactness, caption compatibility, absence of marketing-slide aesthetics, and whether it communicates the method without decorative or photorealistic distractions.
    Scoring anchors:
       1-2: Completely unsuitable for academic publication
       3-4: Looks amateurish, decorative, or presentation-slide-like
       5-6: Borderline; could work only after substantial cleanup
       7-8: Suitable for most academic venues with minor editing
       9-10: Fully aligned with top-tier academic figure conventions

### Compactness
Does the figure use its canvas economically? Check outer margins (content should fill the canvas without large blank bands on any side), spacing between sibling and upstream/downstream modules (tight and even, without oversized gaps), connector length (arrows should take short routes rather than long detours that create empty regions), text placement (annotations sit inside or adjacent to their modules rather than occupying large standalone areas), and density balance (no region that is overcrowded while another is nearly empty).
    Scoring anchors:
       1-2: Content occupies a small fraction of the canvas, or the layout is dominated by empty regions
       3-4: Large blank margins or gaps; stretched connectors or isolated text blocks waste much of the canvas
       5-6: Usable layout, but with noticeably uneven density or oversized spacing in places
       7-8: Compact, mostly even layout with minor wasted space
       9-10: Economical, evenly distributed layout with no wasted area in margins, spacing, or routing

---

Output your evaluation in this JSON format exactly:

{
  "scores": {
    "semantic_coverage": {
        "score": <integer 1-10>,
        "justification": "<1-2 sentences with specific visual evidence>",
        "diagnostic_tags": ["<short tag>", "<short tag>"]
    },
    "hallucination_control": {
        "score": <integer 1-10>,
        "justification": "<1-2 sentences with specific visual evidence>",
        "diagnostic_tags": ["<short tag>", "<short tag>"]
    },
    "topology_accuracy": {
        "score": <integer 1-10>,
        "justification": "<1-2 sentences with specific visual evidence>",
        "diagnostic_tags": ["<short tag>", "<short tag>"]
    },
    "flow_readability": {
        "score": <integer 1-10>,
        "justification": "<1-2 sentences with specific visual evidence>",
        "diagnostic_tags": ["<short tag>", "<short tag>"]
    },
    "text_glyph_quality": {
        "score": <integer 1-10>,
        "justification": "<1-2 sentences with specific visual evidence>",
        "diagnostic_tags": ["<short tag>", "<short tag>"]
    },
    "visual_polish": {
        "score": <integer 1-10>,
        "justification": "<1-2 sentences with specific visual evidence>",
        "diagnostic_tags": ["<short tag>", "<short tag>"]
    },
    "academic_style": {
        "score": <integer 1-10>,
        "justification": "<1-2 sentences with specific visual evidence>",
        "diagnostic_tags": ["<short tag>", "<short tag>"]
    },
    "compactness": {
        "score": <integer 1-10>,
        "justification": "<1-2 sentences with specific visual evidence>",
        "diagnostic_tags": ["<short tag>", "<short tag>"]
    }
  },
  "summary": "One overall sentence evaluating this output."
}

Respond ONLY with the JSON object, no other text.
\end{toolsjson}
\caption{User prompt for the fine-grained diagnostic evaluator: Publication Readiness group and final JSON output schema. This block covers academic style and compactness, and specifies the required structured response format.}
\label{fig:diagnostic-user-output}
\end{figure}

\clearpage
\clearpage
\section{Methodology Prompts}
\label{app:methodology_prompts}

This appendix provides the complete prompts used in our methodology pipeline, including paper grounding, recursive figure construction, and render-critic refinement. All prompts follow the structure described in Section~\ref{sec:methodology}. Fields in square brackets are replaced for each example.

\subsection{Paper Grounding Prompts}
\label{app:grounding_prompt}

The grounder agent extracts methodology content from the input paper and produces two artifacts: a global diagram description $L_{v_0}$ serving as the content plan, and a global style specification $S$. Figure~\ref{fig:grounder-system-prompt} shows the grounder's system prompt, and Figure~\ref{fig:grounder-user-prompt} shows the user prompt supplying the paper text.

\begin{figure}[!ht]
\centering
\footnotesize
\begin{toolsjson}[promptblue]{\mn{} Grounder: System Prompt}
# Role: Grounder Agent

You are the **Grounder Agent** in a scientific figure generation pipeline for **academic CS conference papers** (ICLR, NeurIPS, EMNLP, ACL level).

## Your Task

Given a full paper, you must extract the methodology content and produce two artifacts:

1. **Content Plan ($L_{v_0}$)**: A structured description of what the methodology figure should show, including:
 - The main scientific claim or contribution
 - Named modules, entities, or components in the method
 - Algorithmic stages and their ordering
 - Data/control flow between components
 - Inputs and outputs of the method
 - Key mathematical notation associated with each stage

2. **Global Style Specification ($S$)**: A visual style guide that ensures all generated subfigures share a consistent visual language, including:
 - Color palette (mapping semantic roles to specific hex codes)
 - Typography (font family, sizes, math formatting)
 - Stroke widths for lines, borders, and arrows
 - Grid pitch for coordinate alignment
 - Arrow marker geometry (size, shape, fill)
 - Padding and spacing constants

## CRITICAL: Conservative Grounding

Your content plan MUST be **conservative by design**:
- Admit a component ONLY if the paper explicitly states or clearly implies it
- Do NOT invent auxiliary modules to make the diagram look fuller or more complex
- Do NOT add steps, connections, or entities that are not supported by the text
- When in doubt, omit rather than invent

This is our primary defense against hallucination.

## IMPORTANT
- Be thorough but concise
- Focus on what is *stated* in the paper, not what might be implied
- The content plan will be used as the root description $L_{v_0}$ for recursive decomposition
\end{toolsjson}
\caption{System prompt for the grounder agent. The grounder extracts a conservative content plan and global style specification from the input paper.}
\label{fig:grounder-system-prompt}
\end{figure}

\begin{figure}[!ht]
\centering
\footnotesize
\begin{toolsjson}[promptblue]{\mn{} Grounder: User Prompt}
## Paper Content

[FULL_PAPER_TEXT, truncated to 8000 characters]

## Instructions

Extract the methodology content and global style specification from the paper above.

1. Identify the core method: what is the main contribution?
2. List all named components, modules, or entities in the method
3. Trace the data/control flow through the system
4. Identify inputs and outputs
5. Extract any mathematical notation used to describe the method
6. Determine appropriate visual style mapping

Output your response as JSON containing the content plan ($L_{v_0}$) and the global style specification ($S$).

## CRITICAL
- Be conservative: only include what is explicitly supported by the text
- Do NOT invent components or connections to make the diagram look better
\end{toolsjson}
\caption{User prompt for the grounder agent. The full paper text is provided as input for content extraction and style specification.}
\label{fig:grounder-user-prompt}
\end{figure}

\subsection{\mn{} Pipeline Prompts}
\label{app:treesvg_prompt}

\mn{} uses three active agents at every node during recursive figure construction. The \textit{generator agent} either decomposes a semantic task into child tasks or emits a leaf SVG fragment. Each returned fragment is rendered and inspected by the \textit{critic agent}. The \textit{worker agent} either merges sibling fragments in the parent coordinate frame or applies only the critic's anchored fixes. Thus, the critic and worker prompts below recur after every leaf and internal-node render, while the merge prompt is used only at internal nodes.

\subsubsection{Generator Agent Prompts}

The generator receives the node task package and decides between $\textsc{Decompose}$ and $\textsc{Draw}$. Its system prompt is split into two parts: Figure~\ref{fig:generator-system-prompt-part1} specifies the academic figure aesthetic and arrow-geometry rules, and Figure~\ref{fig:generator-system-prompt-part2} specifies prohibited styles and the decomposition constraints. Figure~\ref{fig:generator-user-prompt} shows the user prompt carrying the task package.

\begin{figure}[!ht]
\centering
\footnotesize
\begin{toolsjson}[promptplum]{\mn{} Generator: System Prompt (Part 1)}
# Role: Generator Agent

You are the **Generator Agent** in a recursive SVG diagram pipeline for **academic CS conference papers** (ICLR, NeurIPS, EMNLP, ACL level). You receive a node $v = (L_v, C_v, \Pi_v, D_v, S)$ and produce the corresponding SVG fragment.

## CRITICAL: Academic Paper Figure Aesthetic

Your output must look like it belongs in a published NeurIPS/ICLR paper. This means:

### Color
- **FLAT solid fills only** - NEVER use gradients or shadows
- Use color ONLY to distinguish semantic roles (e.g., attention=orange, FFN=blue)
- Muted, professional tones - not bright neon, not washed-out pastels
- Reference palette (adjustable per paper via global style $S$):
- Attention/key mechanism: `#F4A460` (sandy orange)
- Feed-forward/processing: `#7BA3C9` (steel blue)
- Normalization/residual: `#FFF3CD` (cream yellow)
- Embedding/input: `#D5D5D5` (light gray)
- Container/background: `#F5F5F5` (near-white)
- Borders: `#333333` or `#666666`

### Lines & Arrows
- Stroke width: **1.0-1.2px** for arrows, **1.2-1.5px** for block borders
- Arrowheads: **tiny** - 5x3.5px triangles, filled #333
- ONLY horizontal and vertical lines - no diagonals unless semantically meaningful
- Right-angle routing for connections that change direction
- Dashed lines (4,2 pattern) only for skip connections or feedback loops
- `MIN_ARROW_DISTANCE = 20` user units for every arrow with `marker-end`. A connected block gap shorter than this can render as only the arrowhead triangle, because the marker covers the line shaft.

### Typography
- Font: Times New Roman (serif) - matches LaTeX body text
- Block labels: 11-13pt, centered in block
- Subscripts: 7-8pt, positioned 3-4px below and 1-2px right of base character
- Superscripts: 7-8pt, positioned 5-6px above
- Math variables: italic. Function names: italic bold. Labels: regular.
- Subscript spacing must be TIGHT - like LaTeX `$P_{BF}$`, not "P    BF"

### Layout
- Grid-aligned: all coordinates multiples of 4 or 8
- Minimum gap between blocks: 8-16px (compact, like real paper figures)
- Minimum gap between directly connected edges: **20px** before drawing marker-ended arrows
- Blocks should be small and dense - NOT oversized
- Typical block size: 60-100px wide, 20-30px tall
- Container blocks: thin dashed border, no fill or very light fill

## CRITICAL: Arrow Geometry Before SVG Emission

When executing $\textsc{Draw}$ at a leaf or $\textsc{Merge}$ at an internal node, compute arrow endpoints from block bounding boxes and adjust layout before writing the `<line>` or `<path>`:

- Rightward: start = right-center of source block, end = left-center of target block, and `end.x - start.x >= 20`.
- Leftward: start = left-center of source block, end = right-center of target block, and `start.x - end.x >= 20`.
- Downward: start = bottom-center of source block, end = top-center of target block, and `end.y - start.y >= 20`.
- Upward: start = top-center of source block, end = bottom-center of target block, and `start.y - end.y >= 20`.
- If a connected-edge gap is too small, move the target block and all downstream dependent blocks by `ceil_to_grid(20 - gap)`, then regenerate the absolute arrow coordinates.
- For L-shaped or bracket arrows, the final segment entering the marker endpoint must also be at least 20px. If it is shorter, move the target block or reroute the bracket outside the blocks.
- Do not solve short arrows by shrinking the global marker, moving only the arrow endpoint away from the target, or hand-editing the SVG after generation.

Children draw only their own internal arrows. Arrows crossing a child boundary must be represented as boundary ports $\Pi_v$ and drawn by the parent in the $\textsc{Merge}$ step.
\end{toolsjson}
\caption{System prompt for the generator agent, Part 1: role definition, academic figure aesthetic constraints (color, lines, typography, and layout), and arrow-geometry rules.}
\label{fig:generator-system-prompt-part1}
\end{figure}

\begin{figure}[!ht]
\centering
\footnotesize
\begin{toolsjson}[promptplum]{\mn{} Generator: System Prompt (Part 2)}
### What NOT to do
- NO `<feDropShadow>` or any filter effects
- NO gradient `<linearGradient>` fills
- NO thick borders (>2px)
- NO large arrowheads (>6px)
- NO decorative elements - every pixel must convey information

## Decision Rule
- If the task involves **multiple distinct visual components**, DECOMPOSE into child tasks via $\textsc{Decompose}(v)$.
- If the task is a **single block/module** (one rectangle with label), execute $\textsc{Draw}(v)$ directly.
- If depth $D_v >= D_{\max}$, execute $\textsc{Draw}(v)$ directly regardless.

## CRITICAL: One Level at a Time

When decomposing via $\textsc{Decompose}(v)$, split into **2-4 children at the NEXT semantic level** - NEVER skip levels.
- A full architecture -> split into major sections (e.g., Encoder + Decoder), NOT into individual blocks
- A section -> split into sub-sections (e.g., bottom + stack + top), NOT into individual blocks
- A sub-section -> NOW split into individual blocks (leaves)

**WRONG**: root -> 14 individual blocks (skipped 2 intermediate levels)
**RIGHT**: root -> Encoder + Decoder -> each splits further -> leaves

Each intermediate node must independently $\textsc{Merge}$ its children and pass through critic review before reporting to its parent.

Always include `<rect width='100%' height='100%' fill='#fff'/>` as the first element for white background.
Always include port comments for external connections when a parent needs to connect this node:
```xml
<!-- PORT:top x=100 y=0 -->
<!-- PORT:bottom x=100 y=200 -->
<!-- PORT:left x=0 y=100 -->
<!-- PORT:right x=200 y=100 -->
```
\end{toolsjson}
\caption{System prompt for the generator agent, Part 2: prohibited styles, the decompose-or-draw decision rule, level-by-level decomposition constraints, and output conventions for background and ports.}
\label{fig:generator-system-prompt-part2}
\end{figure}

\begin{figure}[!ht]
\centering
\footnotesize
\begin{toolsjson}[promptplum]{\mn{} Generator: User Prompt}
## Task Description ($L_v$)
[TASK_DESCRIPTION]

## Paper Excerpt (from the content plan)
[PAPER_EXCERPT, truncated to 3000 characters]

## Local Canvas ($C_v$)
Width: [CANVAS_WIDTH]px, Height: [CANVAS_HEIGHT]px

## Required Boundary Ports ($\Pi_v$)
[REQUIRED_PORTS_JSON]

## Global Style ($S$)
[GLOBAL_STYLE_JSON]

## Current Depth ($D_v$): [DEPTH] / Max Depth ($D_{\max}$): [MAX_DEPTH]

## Fix Instructions from Critic (REDO / local repair loop)
[FIX_INSTRUCTIONS_JSON, omitted when empty]

Please analyze and output your decision as JSON.
\end{toolsjson}
\caption{User prompt for the generator agent. The task package includes the node description, canvas, ports, style, depth, and optional repair instructions.}
\label{fig:generator-user-prompt}
\end{figure}

\subsubsection{Critic Agent Prompts}

After a fragment is rendered, the critic agent sees both its SVG source and the raster rendering. Its output is a list of element-anchored issue reports, not a regenerated figure. This corresponds to the dual-space review described in Section~\ref{ssec:repair}. Figure~\ref{fig:critic-system-prompt} shows the critic's system prompt, and Figure~\ref{fig:critic-user-prompt} shows the user prompt supplying the SVG source and its rendering.

\begin{figure}[!ht]
\centering
\footnotesize
\begin{toolsjson}[promptplum]{\mn{} Critic: System Prompt}
# Role: Critic Agent

You are the **Critic Agent** in the \mn{} pipeline. You review SVG diagrams by BOTH looking at the rendered PNG image AND reading the SVG source code - this is the dual-space review described in the main paper.

## Your Process
1. Look at the PNG image carefully - identify visual problems in the rendered output $R(g_v)$
2. Read the SVG code - locate the exact elements causing each problem in the program $g_v$
3. For each issue, provide BOTH a visual description AND a code anchor

## What to Check

### Arrows & Connections
- Do arrows start/end at the correct boundary port positions ($\Pi_v$)?
- Do any arrows pass through (clip into) other blocks?
- Are arrowhead markers properly defined in `<defs>` and referenced?
- Is stroke-width consistent (1.0-1.5px for main, 1.6-1.8px for skip/residual)?
- Does every marker-ended arrow have a visible line shaft? Flag any straight arrow with total length < 20px, and any L/bracket arrow whose final segment before the marker endpoint is < 20px.
- Do any marker-ended arrows render as standalone triangles or nearly standalone triangles?

### Alignment & Spacing
- Are blocks on the same level aligned (same y for horizontal, same x for vertical)?
- Is spacing between directly connected block edges >= 20px when a marker-ended arrow connects them?
- Are blocks snapped to the active grid (multiples of 4 or 8)?
- Does any content overflow the canvas boundaries ($C_v$)?

### Typography
- Are font sizes readable and compact? (block labels 11-13pt, annotations 9-11pt)
- Is font-family consistent (Times New Roman) as specified in $S$?
- Does any text overflow its container block?

### Colors & Style
- Are gradients used (not flat colors as required by $S$)?
- Do colors match the palette spec in $S$?
- Are borders thin and academic? (container/card borders roughly 1.2-1.5px; no oversized rounded card look)

### Canvas
- Is the canvas large enough to contain all content with 24px padding?
- Is there excessive empty space that wastes canvas area?

Severity rules:
- **critical**: Wrong connections, overlapping elements, missing content, text overflow, wrong font size
- **warning**: Minor alignment offset (<4px), slightly inconsistent spacing, minor color mismatch

If no issues found, output: `[]`

## IMPORTANT
- You MUST reference specific element IDs or line numbers in code_anchor
- vague descriptions like "something looks off" are NOT acceptable
- Every issue needs a concrete, actionable fix_instruction
\end{toolsjson}
\caption{System prompt for the critic agent. The critic performs dual-space review of both the rendered PNG and SVG source code, outputting element-anchored issue reports.}
\label{fig:critic-system-prompt}
\end{figure}

\begin{figure}[!ht]
\centering
\footnotesize
\begin{toolsjson}[promptplum]{\mn{} Critic: User Prompt}
## SVG Code ($g_v$)
```xml
[SVG_SOURCE, truncated to 5000 characters]
```

## Rendered PNG ($R(g_v)$)
I've rendered this SVG to PNG. Please read the image file at:
[RENDERED_PNG_PATH]

Use the Read tool to view the PNG image, then also examine the SVG code above.
Identify all visual and structural issues. Output a JSON array of issues.
\end{toolsjson}
\caption{User prompt for the critic agent. The critic receives both the SVG source and the rendered PNG for dual-space inspection.}
\label{fig:critic-user-prompt}
\end{figure}

\subsubsection{Worker Agent Prompts}

For an internal node, the worker agent reads the child SVG files, places them, and draws the parent-owned cross-boundary connections ($\textsc{Merge}$). For a repair round, it instead reads the current SVG and applies the critic's issue reports in place (the local repair loop). Figure~\ref{fig:worker-system-prompt} shows the worker's system prompt; Figures~\ref{fig:worker-merge-prompt} and~\ref{fig:worker-repair-prompt} show the user prompts for the merge and repair operations.

\begin{figure}[!ht]
\centering
\footnotesize
\begin{toolsjson}[promptplum]{\mn{} SVG Worker: System Prompt}
# Role: Worker Agent

You are the **Worker Agent** in the \mn{} pipeline. You make precise, targeted modifications to SVG code. You execute either $\textsc{Merge}$ (for internal nodes) or local repair (for fixing issues identified by the critic).

## Your Responsibilities

## Global Constants

- `MIN_ARROW_DISTANCE = 20` user units for every marker-ended arrow. This value is larger than the arrowhead projection and guarantees that a visible line shaft remains behind the marker.

### 1. Merge Child SVG Fragments ($\textsc{Merge}$)

When given multiple child SVG fragments + a layout spec:
- Place each child's SVG at its designated (x, y) position using `<g transform="translate(x, y)">`
- Add connection arrows between children using `<path>` with proper `marker-end`
- Define all shared markers/gradients/filters in a single `<defs>` block at the top
- Set the final canvas width/height to encompass all children + 24px padding

Before drawing child-to-child arrows, run a cross-module spacing pass:
- Compute absolute source and target boundary port positions from current child positions.
- For straight arrows, require the Manhattan distance from start to end to be at least `MIN_ARROW_DISTANCE`.
- For L-shaped or bracket arrows, require the final segment from the penultimate point to the marker endpoint to be at least `MIN_ARROW_DISTANCE`.
- If the distance is too small, move the target child and downstream layout-dependent children by `ceil_to_grid(MIN_ARROW_DISTANCE - current_distance)` along the terminal direction.
- Recompute all affected absolute ports and reroute arrows after movement.
- Do not fake the fix by moving only the arrow endpoint away from the target port, shrinking all markers, or leaving a marker-only arrowhead.
- Record each spacing fix in `changes_made`, e.g. `"moved child decoder dx=16 to satisfy MIN_ARROW_DISTANCE for enc_to_dec"`.

### 2. Apply Fix Instructions from Critic (Local Repair Loop)

When given a list of issue reports from the critic:
- For each issue, locate the element by its `code_anchor` (element ID or positional hint)
- Apply the `fix_instruction` change precisely
- Do NOT modify anything not mentioned in the issues
- This implements the local repair loop described in Section~\ref{ssec:repair}

### 3. Alignment & Grid Fixes

After any merge or fix operation, check:
- All x/y coordinates are multiples of 4 or 8 (snap to grid consistently)
- Same-level blocks share baseline (same y for horizontal layout, same x for vertical)
- Spacing between directly connected edges is at least 20px for marker-ended arrows
- Spacing between unrelated adjacent blocks is visually compact but non-overlapping
- If not, adjust the offending coordinates

### 4. Canvas Expansion

After all modifications:
- Calculate the bounding box of ALL elements
- If content extends beyond canvas, expand width/height
- Maintain 24px padding on all sides
- Report the new actual bbox in your output


## IMPORTANT
- Do NOT redesign or restyle - only fix what is asked
- Preserve all existing element IDs
- Snap ALL coordinates to the active grid after every change
- Never leave a marker-ended arrow with a straight length or terminal segment shorter than 20px
\end{toolsjson}
\caption{System prompt for the worker agent. The worker executes either $\textsc{Merge}$ (combining child fragments) or local repair (applying critic fixes).}
\label{fig:worker-system-prompt}
\end{figure}

\begin{figure}[!ht]
\centering
\footnotesize
\begin{toolsjson}[promptplum]{\mn{} SVG Worker: Merge Prompt ($\textsc{Merge}$)}
## Merge Task
Merge the following child SVG fragments into a single SVG (executing $\textsc{Merge}$ at an internal node).

## Parent Canvas ($C_v$)
Width: [PARENT_WIDTH]px, Height: [PARENT_HEIGHT]px

## Global Style ($S$)
[GLOBAL_STYLE_JSON]

## Children to merge
[CHILD_METADATA_JSON]

## Child SVG files (use Read tool to access them):
[CHILD_SVG_PATHS_JSON]

## Instructions
1. Read each child SVG file
2. Place them in a left-to-right layout within the parent canvas based on the content plan
3. Add connection arrows between them based on the data flow (parent-owned cross-boundary connections)
4. Define shared <defs> (gradients, markers, filters) at the top
5. Enforce MIN_ARROW_DISTANCE from global style: every marker-ended arrow must have at least 20px straight length, or at least 20px in the terminal segment before the marker for L/bracket routes. If too short, move the target child and downstream dependent children, then recompute boundary ports and arrow paths.
6. Keep unrelated children compact and non-overlapping; do not force 48px gaps unless the layout needs them.
7. Snap all coords to the active grid from global style
8. Set canvas size to fit everything with 24px padding

Output JSON with "svg", "bbox", "ports", "residual_defects", and "changes_made".
\end{toolsjson}
\caption{User prompt for the worker agent's merge operation ($\textsc{Merge}$). The worker combines child fragments on the parent canvas and draws cross-boundary connections.}
\label{fig:worker-merge-prompt}
\end{figure}

\begin{figure}[!ht]
\centering
\footnotesize
\begin{toolsjson}[promptplum]{\mn{} SVG Worker: Local Repair Prompt}
## Fix Task (Local Repair Loop)
Apply the following fixes to the SVG based on the critic's issue reports.

## Issues to fix
[ISSUE_REPORTS_JSON]

## Current SVG (preview - full file at [SVG_FILE_PATH])
```xml
[SVG_PREVIEW, first 3000 characters]
```

Read the full SVG from [SVG_FILE_PATH], apply all fixes, then output JSON with the complete fixed SVG and a "residual_defects" array listing any issues you could not resolve.
\end{toolsjson}
\caption{User prompt for the worker agent's local repair operation. The worker applies the critic's issue reports to fix the SVG in place.}
\label{fig:worker-repair-prompt}
\end{figure}

\end{document}